# Edge Detection: A Collection of Pixel based Approach for Colored Images

B O. Sadiq
Department of Electrical and Computer Engineering,
Ahmadu Bello University, Zaria

S.M. Sani
Department of Electrical and Computer Engineering,
Ahmadu Bello University, Zaria

S. Garba
Department of Electrical and Computer Engineering,
Ahmadu Bello University, Zaria

## ABSTRACT
The existing traditional edge detection algorithms process a single pixel on an image at a time, thereby calculating a value which shows the edge magnitude of the pixel and the edge orientation. Most of these existing algorithms convert the coloured images into gray scale before detection of edges. However, this process leads to inaccurate precision of recognized edges, thus producing false and broken edges in the image. This paper presents a profile modelling scheme for collection of pixels based on the step and ramp edges, with a view to reducing the false and broken edges present in the image. The collection of pixel scheme generated is used with the Vector Order Statistics to reduce the imprecision of recognized edges when converting from coloured to gray scale images. The Pratt Figure of Merit (PFOM) is used as a quantitative comparison between the existing traditional edge detection algorithm and the developed algorithm as a means of validation. The PFOM value obtained for the developed algorithm is 0.8480, which showed an improvement over the existing traditional edge detection algorithms.

## General Terms
Coloured Images and Pratt Figure of Merit

## 1. INTRODUCTION
An edge in an image can be defined as a single pixel with local discontinuity in intensity [1]. Edge detection is a process of identifying these local discontinuities in images using an algorithm [2]. Various factors are responsible for these discontinuities in images, which are not limited to light, shadows and illumination [3]. These discontinuities produced four different types of edge profiles which are the step, ramp, roof and ridge edges. There exists the traditional edge detection algorithms which are the Sobel Edge Detection Algorithm, Prewitt Edge Detection Algorithm, Roberts Edge Detection Algorithm, Laplacian Edge Detection Algorithm and the Canny Edge Detection Algorithm. This existing algorithms processes a single pixel on an image at a time, thereby calculating a value which shows the edge magnitude of the pixel and the edge orientation [4]. These existing traditional edge detection algorithms could not perform effectively on coloured images unless converted to gray scale [5]. The conversion of coloured images to gray scale before detection of edges often leads to inaccurate precision of recognized edges, thus producing false and broken edges in the image. With recent advancement in technology and the increased usage of coloured images, there is need to develop an effective edge detection algorithm for coloured images without converting to gray scale [6]. Numerous researchers have developed different algorithms for edge detection in images such as [7], [8] and [9] with a view to reducing the shortcomings of the existing traditional edge detection algorithms. The authors in [7] developed an algorithm for coloured images using a Sobel mask. The algorithm processes a single pixel on an image thereby using a thresholding scheme to detect a pixel as an edge. But this technique lead to misplaced edges. In [8], a maximum directional difference of the sum of gray values when each component of the image taken separately were calculated for each pixel. A threshold value is then selected with a view to generating the edges. A transformation technique was used to combine the individual gray values to produce the edge map. However, missing edges exists in the generated output edge map. The authors in [9] presented an approach using the ant colony optimization technique. But the images are converted to gray scale before detection of edges. This leads to false and broken edges.

In view of the shortcomings associated with the existing traditional edge detection algorithms, a collection of pixels based approach using vector order statistics for coloured images is presented with a view to reducing the false and broken edges that exists in generated edge maps.

## 2. VECTOR ORDER STATISTICS
Coloured images are 3-D arrays that assign three numerical values to each pixel, each value corresponding to the red, green and blue (RGB) image channel component respectively. While, Gray-scale images are 2-D arrays that assign one numerical value to each pixel which is the representative of the intensity at this point. Gray scale images are generated from coloured images by suppressing the RGB component to 1 and 0. The value 1 is the white region and the value black is the dark region [10]. Since coloured images consists of three channels, the ordering of the pixels in each of the RGB component in a vector form is known as the vector order statistics [11]. Coloured images consists of 3 channels which are 8-bits each, denoted by R-channel, G-channel and B-channel. During processing of coloured images, the RGB components in the image is viewed as a vector field that maps a point in the image plane to a three-dimensional (3-D) vector in the colour space. In the three dimensional space, the pixel value is the vector M= (R, G, B), where M is a function of (j, k) in the image. The vector M can be represented at any point using [12] equation 2.1

$$F = \begin{bmatrix} R_j & R_k \\ G_j & G_k \\ B_j & B_k \end{bmatrix} = (M_j, M_k) \qquad (2.1)$$

## 3. METHODOLOGY

i. Develop a pixel collection scheme for coloured images.

ii. Apply the generated collecton scheme to vector order statistics

iii. Generate the output edge map based on the threshold value

### 3.1 Collection Scheme

A collection scheme is developed for the algorithm with a view to minimizing the effect of false and broken edges in the image. The collection scheme are based on the step and ramp edges. The collection of set of pixels are depicted in Figure 1 and 2

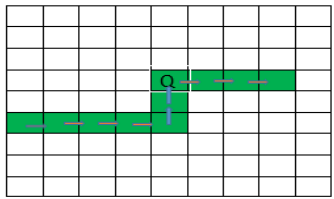

**Figure 1: Collection of Pixel based on Step Edge**

The Figure 2 shows the collection scheme based on Roof edge

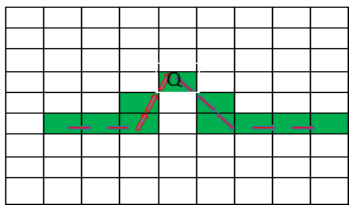

**Figure 2: Collection of Pixel based on Roof Edge**

These collections of set of pixels are then used to generate a collection scheme which are then used as a mask. The collection scheme for both the step and ramp edges are shown in Figure 3 and 4

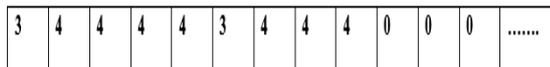

**Figure 3: Collection Scheme Based on Step Edge**

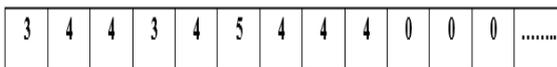

**Figure 4: Collection Scheme Based on Roof Edge**

The developed collection scheme is used to process a collection of pixels instead of the single pixel value as used by the existing traditional edge detection algorithm. The collection scheme are used to generate two mask which are applied to the image with a view to finding the gradient. The generated mask are depicted in Figure 5

| 3 | 4 | 4 |   | 3 | 4 | 4 |
|---|---|---|---|---|---|---|
| 3 | 4 | 5 |   | 4 | 4 | 3 |
| 4 | 4 | 4 |   | 4 | 4 | 4 |
| $F_x$ |  |  |   | $F_y$ |  |  |

**Figure 5: Mask Used to Find the Image Gradient**

### 3.2 Vector Approach and Algorithm

i. From the computer database, choose the image as the input

ii. From the input image, generate a 3x3 pixel window

iii. From each pixel in the window, a vector of size 3 is used to describe the colour. This is written as $P_{p,q}$ RGB. The vector is the RGB values of that pixel.

iv. Determine the Euclidean distance between a given $P_{p,q}$ RGB and all other $P_{p,q}$ RGB in the window. This generates a new set of scalars $A_0$ to $A_8$

v. Calculate the vector range finding the difference between the first and last pixel resulting from the ordering.

vi. Suppress the edges in the images using the developed collection scheme

vii. Set a user threshold value to determine if a pixel is an edge or not.

The flowchart of the algorithm is depicted in Figure 6

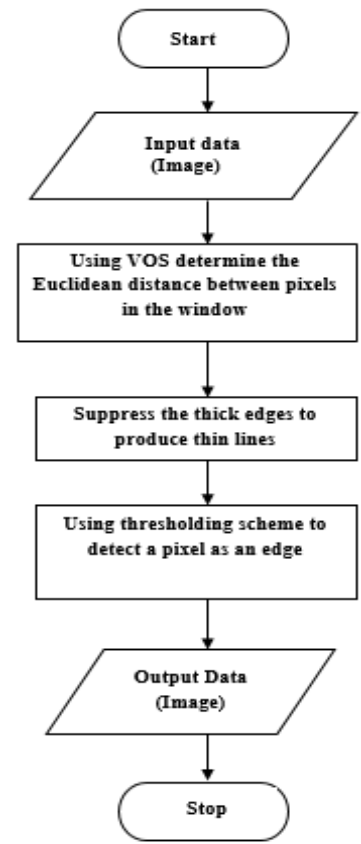

**Figure 6: Flow Chart of the Algorithm**

The input image is depicted in Figure 7

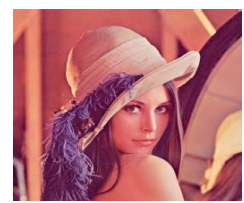

**Figure 7: Input image**

## 3.3 Non-Maximum Suppression

The non-maximum suppression is used to suppress the thick edges produced by the vector order statistics with a view to achieving thin and continuous edge lines. The steps used to achieve this are as follows:

i. Splitting the image into four directions using an 8 neighboring pixels.

ii. Check the left and the right neighboring gradient values.

iii. If the gradient at the present pixel is greater than its left and right neighbors, then it is considered an edge, else it is discarded.

## 4. RESULTS AND DISCUSSIONS

The Pratt Figure of Merit (PFOM) is a method used to provide a quantitative comparison between edge detection algorithms in image processing [13]. The PFOM is determined by a mathematical expression as in equation 4.1. the PFOM measures the value of detected edges between 0 and 1. As the value get closer to 1, it shows better detected edge values [14].

$$R = \frac{1}{Max(N_I, N_A)} \sum_{k=1}^{N_A} \frac{1}{1 + md^2(k)} \quad (4.1)$$

Where: $N_I$ is the number of actual edges

$N_A$ is the number of detected edges

m is a scaling constant set to 1/9.

d (k) denotes the distance from the actual edge to the corresponding detected edge

The output of applying the proposed edge detection algorithm to the input image and the traditional edge detection algorithms are depicted in Figure 8.

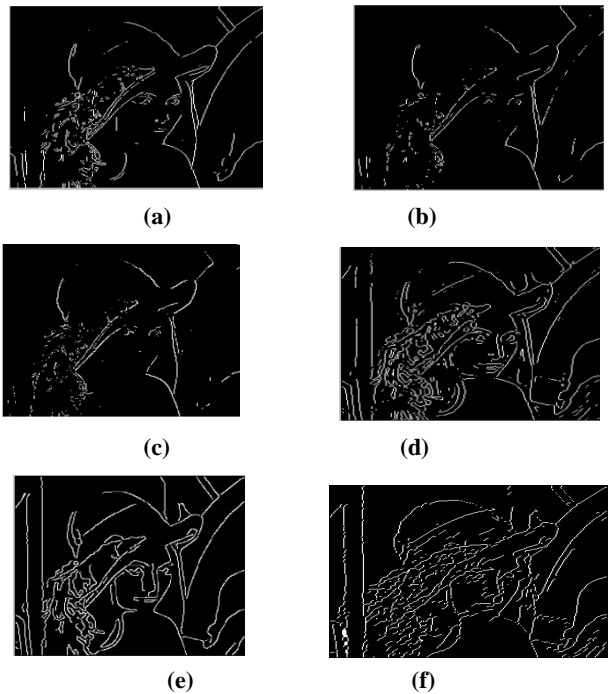

(a)　(b)

(c)　(d)

(e)　(f)

**Figure 8: Output Result of Applying the Proposed Edge Detection Algorithm**

The output result presented in Figure 8 is that of the existing traditional edge detection algorithms and the proposed edge detection algorithms. The Figures (a), (b), (c), (d), (e) and (f) are the output result of the Sobel, Prewitt, Roberts, Laplacian, Canny and the proposed edge detection algorithms respectively. In comparison with the existing traditional edge detection algorithms, the table 1 shows the result obtained using the PFOM

**Table 1: Comparison of Traditional Edge Detection Algorithm and the Developed Algorithm Using PFOM**

| Edge detection algorithm | PFOM |
|---|---|
| Sobel edge detection algorithm | 0.4209 |
| Prewitt edge detection algorithm | 0.4195 |
| Robert edge detection algorithm | 0.4181 |
| Laplacian edge detection algorithm | 0.7048 |
| Canny edge detection algorithm | 0.8472 |
| Proposed edge detection algorithm | 0.8480 |

The PFOM showed that the output result of the developed algorithm has more detected and localized edges than those of the existing traditional edge detection algorithms. The plot in Figure 9 shows the PFOM for the various edge detection algorithms.

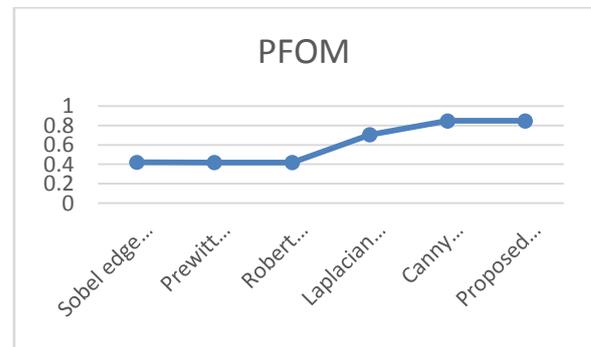

**Figure 9: PFOM for Various Edge Detection Algorithms**

## 5. CONCLUSION

A collection of pixel based approach for edge detection has been proposed with a view to reducing false and broken edges that exists in images. The algorithm developed was based on the vector order statistics with a view to detecting edges for coloured images. The collection scheme was based on the step and ramp edges. The algorithm was developed and implemented in MATLAB 2013b script. Based on the result obtained in visual and quantitative comparison with the traditional edge detection algorithms, it can be concluded that the pixels that constitute the edges in an image have been extracted completely. This result signifies improvement in detection of edges as compared with the existing traditional edge detection algorithms using the Pratt Figure of Merit (PFOM). The developed algorithm can be extended to object detection, tracking and region based segmentation.